# Exploring the Accuracy Potential of IMU Preintegration in Factor Graph Optimization

Hailiang Tang, Xiaoji Niu, Tisheng Zhang, Jing Fan, and Jingnan Liu

*Abstract*—Inertial measurement unit (IMU) preintegration is widely used in factor graph optimization (FGO); *e.g.*, in visual-inertial navigation system and global navigation satellite system/inertial navigation system (GNSS/INS) integration. However, most existing IMU preintegration models ignore the Earth's rotation and lack delicate integration processes, and these limitations severely degrade the INS accuracy. In this study, we construct a refined IMU preintegration model that incorporates the Earth's rotation, and analytically compute the covariance and Jacobian matrix. To mitigate the impact caused by sensors other than IMU in the evaluation system, FGO-based GNSS/INS integration is adopted to quantitatively evaluate the accuracy of the refined preintegration. Compared to a classic filtering-based GNSS/INS integration baseline, the employed FGO-based integration using the refined preintegration yields the same accuracy. In contrast, the existing rough preintegration yields significant accuracy degradation. The performance difference between the refined and rough preintegration models can exceed 200% for an industrial-grade MEMS module and 10% for a consumer-grade MEMS chip. Clearly, the Earth's rotation is the major factor to be considered in IMU preintegration in order to maintain the IMU precision, even for a consumer-grade IMU.

*Index Terms*—IMU preintegration, factor graph optimization, inertial navigation system, GNSS/INS integration.

## I. INTRODUCTION

Inertial measurement units (IMUs) have been employed in navigation systems since the last century [1], [2]. By integrating the angular rates and linear accelerations measured by the IMU, an inertial navigation system (INS) can provide full navigation parameters, including position, velocity, and attitude information [3], [4]. With the rapid development of micro-electro-mechanical systems (MEMS), MEMS IMUs have been widely employed in navigation applications, because of their low cost, low power consumption, and small size. Typically, the IMU is aided by a global navigation satellite system (GNSS) receiver and, hence, a GNSS/INS integrated navigation system is constructed, which can provide a continuous navigation solution. Traditionally, GNSS/INS integration has been based on a Kalman filter framework, which has incorporated an extended Kalman filter (EKF), an unscented Kalman filter [4], or other alternatives. In particular, EKF-based GNSS/INS integration has become a mature technology that constitutes a benchmark for the potential system accuracy, and especially the INS accuracy [1]-[4]. Recently, factor graph optimization (FGO) has been proven to yield better performance and robustness than Kalman filter-based frameworks as regards GNSS/INS integration in GNSS-denied environments [5], [6]; thus, FGO-based GNSS/INS integration has become a research focus.

MEMS IMUs have also been widely used in visual-inertial navigation system (VINS) applications [7]-[21]. Historically, the VINS state-estimation problem has been addressed through filtering, with the IMU measurements being propagated and the visual measurements being used to form updates [7], [8]. Recently, however, FGO has proven to be more accurate and efficient than filtering-based approaches for solving the VINS maximum a posteriori (MAP) estimation [16], [17]. However, it has historically been difficult to incorporate IMUs into FGO because of certain acceleration measurement and bias characteristics [18].

To overcome these problems, the theory of IMU preintegration was proposed by Lupton and Sukkarieh [9], [10], who integrated IMU measurements in a local reference frame and linearized those measurements about the current bias estimations to construct an FGO relative motion constraint factor. Subsequently, Forster *et al.* [11] built upon [10] to develop a preintegration model that properly addressed the manifold structure of the SO(3) rotation group, and seamlessly integrated that preintegration model into a visual-inertial pipeline under the unifying framework of FGO. In a related study [14] IMU on-manifold preintegration was extended to the $SE_2(3)$ Lie group to allow expression of the extended pose (position, velocity, orientation). Besides on-manifold preintegration, quaternion-based preintegration model has also attracted intense attention [12], [13], [15], because of the special form of the quaternion. Moreover, to overcome the discretization effect and improve the accuracy of IMU preintegration, IMU motion integration models in continuous time [18]-[21] have been proposed. For example, Eckenhoff *et al.* [18], [19] proposed a new analytical preintegration model for the IMU kinematics in continuous time; this approach yielded improved accuracy for VINSs. In addition, a switched linear system was used to model IMU motion in continuous time [20], [21], outperforming state-of-the-art IMU preintegration models. However, most preintegration approaches to date have been roughly constructed and have

This research is funded by the National Key Research and Development Program of China (No. 2020YFB0505803), and the National Natural Science Foundation of China (No. 41974024). *(Corresponding authors: Tisheng Zhang; Jingnan Liu.)*

Hailiang Tang, Xiaoji Niu, Tisheng Zhang, Jing Fan, and Jingnan Liu are with the GNSS Research Center, Wuhan University, Wuhan 430079, China (e-mail: {thl, xjniu, zts, jingfan, jnliu}@whu.edu.cn).

neglected important factors, such as the Earth's rotation.

Current MEMS IMUs exhibit notably improved precision with reduced cost, and some industrial-grade MEMS modules have even achieved the close precision compared to tactical-grade IMUs [3]; such modules have been widely adopted in VINS and GNSS/INS integration applications. For example, ADIS16465 [23] from Analog Devices, Inc., (ADI) has an in-run bias instability of 2 °/h (Allan deviation). As the Earth's rotation rate is 15 °/h, ADIS16465 can sense this rotation to some extent, and thus the Earth's rotation is non-negligible for this device. To improve the preintegration accuracy for this kind of high-precision IMU, the Earth's rotation has been considered in recent IMU preintegration models [14], [15]. For example, Barrau *et al.* [14] proposed a preintegration model incorporating the rotating Earth for a high-precision VINS, and formulated the preintegration while considering the centrifugal force and Coriolis effect. Their rigorous treatment of the Coriolis effect was achieved using a nontrivial trick; however, they did not illustrate the accuracy improvement obtained following consideration of the Earth's rotation [14]. In our previous work [15], the Earth's rotation and gravity change were both considered in a preintegration model for visual-inertial odometry. However, experiment results indicated that, although the Earth's rotation can be omitted for MEMS IMUs, it cannot be neglected for navigation-grade IMUs [15]. According to our dedicated analyses, VINS involves various parameter settings and multiple impact factors, which may notably disturb the result and confuse the conclusion. Hence, the impact of the Earth's rotation in preintegration cannot been effectively determined for a MEMS IMU in [15]. Thus, we believe that the Earth's rotation is an essential factor that may significantly influence IMU preintegration accuracy, even for a MEMS IMU.

In this study, to explore the accuracy potential of IMU preintegration, we construct a refined model incorporating the Earth's rotation. To mitigate the impact caused by sensors other than IMU in the evaluation system, we adopt the concise FGO-based GNSS/INS integration, rather than VINS, to quantitatively evaluate the accuracy of the refined preintegration. In addition, simulated GNSS outage, an evaluation standard [24] for GNSS/INS integrated systems, is employed for evaluation. Moreover, as the classic EKF-based GNSS/INS integration is an established benchmark for the potential INS accuracy, it is treated as a baseline. The main contributions of this study are as follows:

● A refined IMU preintegration model incorporating the Earth's rotation is constructed, along with an analytical solution for noise propagation.

● The refined preintegration model is incorporated into an FGO-based GNSS/INS integrated navigation system for accuracy evaluation, with analytical expressions for the measurement residuals.

● Four different MEMS IMUs are adopted to quantitatively evaluate the accuracy of the refined preintegration. Our vehicle GNSS/INS dataset is available online (https://github.com/i2Nav-WHU/awesome-gins-datasets).

## II. PRELIMINARIES

This section briefly introduces the attitude parameterization and coordinate frames used in this study.

### A. Attitude Parameterization

In this study, we use a quaternion to represent the attitude parameter. The general frames "a," "b," and "c" are used throughout this section. The quaternion, rotation vector, and direction cosine matrix (DCM) for the rotation from the "b" to "a" frame are represented by $\mathbf{q}_b^a$, $\boldsymbol{\phi}_b^a$, and $\mathbf{C}_b^a$, respectively. Here, $\mathbf{q}_b^a$ is a four-dimensional vector composed of a scalar component $q_w$ and a three-dimensional vector component $\mathbf{q}_v$, such that

$$\mathbf{q}_b^a \triangleq \begin{bmatrix} q_w \\ \mathbf{q}_v \end{bmatrix} = \begin{bmatrix} q_w & q_x & q_y & q_z \end{bmatrix}^T. \quad (1)$$

Moreover, $\mathbf{q}_b^a$, $\boldsymbol{\phi}_b^a$, and $\mathbf{C}_b^a$ can be expressed in terms of each other [1]-[4], [22]. For example, $\mathbf{q}_b^a$ can be expressed in terms of $\boldsymbol{\phi}_b^a$ as follows:

$$\mathbf{q}_b^a = \begin{bmatrix} \cos \| 0.5 \boldsymbol{\phi}_b^a \| \\ \dfrac{\sin \| 0.5 \boldsymbol{\phi}_b^a \|}{\| 0.5 \boldsymbol{\phi}_b^a \|} 0.5 \boldsymbol{\phi}_b^a \end{bmatrix}. \quad (2)$$

If $\boldsymbol{\phi}_b^a$ represents a small-angle rotation, we have

$$\mathbf{q}_b^a \approx \begin{bmatrix} 1 \\ 0.5 \boldsymbol{\phi}_b^a \end{bmatrix}. \quad (3)$$

In this study, $\otimes$ denotes the quaternion product, and the chain rule of the quaternion product can be expressed as follows:

$$\mathbf{q}_c^a = \mathbf{q}_b^a \otimes \mathbf{q}_c^b = [\mathbf{q}_b^a]_L \mathbf{q}_c^b = [\mathbf{q}_c^b]_R \mathbf{q}_b^a, \quad (4)$$

where $[\mathbf{q}]_L$ and $[\mathbf{q}]_R$ are, respectively, the left and right quaternion product matrixes

$$[\mathbf{q}]_L = q_w \mathbf{I} + \begin{bmatrix} 0 & -\mathbf{q}_v^T \\ \mathbf{q}_v & [\mathbf{q}_v \times] \end{bmatrix}, \quad [\mathbf{q}]_R = q_w \mathbf{I} + \begin{bmatrix} 0 & -\mathbf{q}_v^T \\ \mathbf{q}_v & -[\mathbf{q}_v \times] \end{bmatrix}. \quad (5)$$

Here, $[\mathbf{a} \times]$ denotes the cross-product (skew-symmetric) form of a vector $\mathbf{a} = \begin{bmatrix} a_x & a_y & a_z \end{bmatrix}^T$, and can be defined as follows:

$$[\mathbf{a} \times] = \begin{bmatrix} 0 & -a_z & a_y \\ a_z & 0 & -a_x \\ -a_y & a_x & 0 \end{bmatrix}. \quad (6)$$

The time derivative of $\mathbf{q}_b^a$ is defined as follows:

$$\dot{\mathbf{q}}_b^a = \frac{1}{2} \mathbf{q}_b^a \otimes \begin{bmatrix} 0 \\ \boldsymbol{w}_{ab}^b \end{bmatrix}, \quad (7)$$

where $\boldsymbol{w}_{ab}^b$ represents the angular velocity vector of the "b" frame relative to the "a" frame, with projection in the "b" frame. For more details on the attitude parameterization, readers can refer to [22].

### B. Coordinate Frames

The inertial frame (i-frame) is an ideal frame of reference in which gyroscopes and accelerometers fixed to the i-frame have zero outputs. The origin of the Earth frame (e-frame) is at the center of mass of the Earth, with axes that are fixed with respect to the Earth. The angular velocity vector of the e-frame with respect to the i-frame projected to the e-frame is

$$\boldsymbol{w}_{ie}^e = \begin{bmatrix} 0 & 0 & w_e \end{bmatrix}^T, \quad (8)$$

where $w_e$ is the magnitude of the Earth's rotation rate ($7.2921158 \times 10^{-5}$ rad/s) [3]. The body frame (b-frame) is the frame in which the angular rates and accelerations generated by the strapdown gyroscopes and accelerometers are resolved. The navigation frame (n-frame) is a local geodetic frame with origin coinciding with that of the sensor frame, with its x-axis pointing toward north, its y-axis pointing toward east, and its z-axis completing a right-hand orthogonal frame, *i.e.* the north-east-down (NED) system. Readers may refer to [1]-[4] for more details of the coordinate frames.

The world frame (w-frame) is the reference coordinate frame in the FGO. In this study, the n-frame at the initial position is selected as the w-frame. The gravity vector is assumed to be unchanged in the w-frame and can be defined as

$$\boldsymbol{g}^w = [\,0 \quad 0 \quad g^n(\varphi_0, h_0)\,]^T, \tag{9}$$

where $\varphi_0$ and $h_0$ are the geodetic latitude and geodetic height at the initial position in the WGS-84 ellipsoid model [3], respectively. In the w-frame, the Earth's rotation rate can be expressed as

$$\boldsymbol{w}_{ie}^w = [\,w_e\cos\varphi_0 \quad 0 \quad -w_e\sin\varphi_0\,]^T, \tag{10}$$

where $\boldsymbol{w}_{ie}^w$ is also assumed unchanged in the w-frame.

## III. REFINED IMU PREINTEGRATION MODEL

Inspired by the high-accuracy INS mechanization algorithm [1]-[4], we further refine the IMU preintegration model in [13] to incorporate the Earth's rotation. In this section, the IMU kinematic model is illustrated first, followed by the delicate IMU motion integration and preintegration processes. Then, the noise propagation of the preintegration measurement is determined, along with the bias-update procedure.

### A. Kinematic Model

An IMU can measure angular rates $\tilde{\boldsymbol{w}}_{ib}^b$ and accelerations (specific force) $\tilde{\boldsymbol{f}}^b$. IMU measurements are affected by various errors, including biases, scale factors, non-orthogonalities, and white noise. In this study, we only consider additive noise $\boldsymbol{n}$ and slowly varying biases $\boldsymbol{b}$:

$$\tilde{\boldsymbol{w}}_{ib}^b = \boldsymbol{w}_{ib}^b + \boldsymbol{b}_g + \boldsymbol{n}_g,$$
$$\tilde{\boldsymbol{f}}^b = \boldsymbol{f}^b + \boldsymbol{b}_a + \boldsymbol{n}_a, \tag{11}$$

where $\boldsymbol{b}_g$ and $\boldsymbol{b}_a$ represent the gyroscope and accelerometer bias errors, respectively, and $\boldsymbol{n}_g$ and $\boldsymbol{n}_a$ represent the gyroscope and accelerometer white noise, respectively.

Based on the classic high-accuracy INS kinematic model [1]-[4], we omit certain tiny terms and obtain the following reduced model:

$$\dot{\boldsymbol{p}}_{wb}^w = \boldsymbol{v}_{wb}^w,$$
$$\dot{\boldsymbol{v}}_{wb}^w = \mathbf{C}_b^w \boldsymbol{f}^b + \boldsymbol{g}^w - 2[\boldsymbol{w}_{ie}^w \times]\boldsymbol{v}_{wb}^w,$$
$$\dot{\mathbf{q}}_b^w = \frac{1}{2}\mathbf{q}_b^w \otimes \begin{bmatrix} 0 \\ \boldsymbol{w}_{wb}^b \end{bmatrix}, \; \boldsymbol{w}_{wb}^b = \boldsymbol{w}_{ib}^b - \mathbf{C}_w^b \boldsymbol{w}_{ie}^w, \tag{12}$$

where $\boldsymbol{w}_{ie}^w$ and $\boldsymbol{g}^w$ are the Earth's rotation rate and gravity vector in the w-frame, respectively (see (9) and (10)). If we omit $\boldsymbol{w}_{ie}^w$ from (12), the kinematic model degenerates to an extremely rough version, as in [13]. The Coriolis acceleration $2[\boldsymbol{w}_{ie}^w \times]\boldsymbol{v}_{wb}^w$ due to the Earth's rotation is maintained to improve the model accuracy. For further reduction of the motion model, readers may refer to [3] and [27].

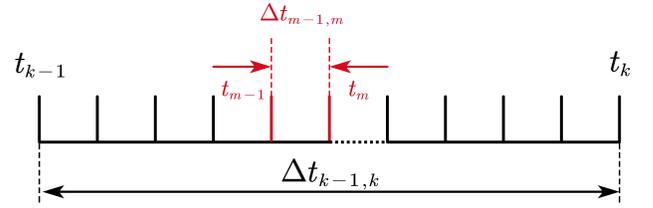

Fig. 1. IMU sample time and preintegration intervals. The preintegration interval is from $t_{k-1}$ to $t_k$. The interval $\Delta t_{m-1,m} = t_m - t_{m-1}$ is an IMU sample time interval.

### B. Motion Integration

During integration interval $[t_{k-1}, t_k]$, the duration can be computed by $\Delta t_{k-1,k} = t_k - t_{k-1}$. Here, times $t_{m-1}$ and $t_m$ are consecutive IMU sample time instants in the interval, as showed in Fig. 1. The incremental angles $\Delta\boldsymbol{\theta}_m$ and incremental velocities $\Delta\boldsymbol{v}_{f,m}$ can be computed by integrating the angular rates and accelerations, or acquired directly from the IMU (some IMUs provide the incremental measurements). In this section, the IMU measurements are compensated with the estimated biases in advance, although they are not explicitly expressed in the formulation (Sections III.D and III.E).

The IMU motion integration can be formulated considering the kinematic model in (12). The attitude update from $t_{m-1}$ to $t_m$ can be expressed as follows:

$$\mathbf{q}_{b_m}^w = \mathbf{q}_{w_{i(m-1)}}^w(t_m) \otimes \mathbf{q}_{b_{i(m-1)}}^{w_{i(m-1)}} \otimes \mathbf{q}_{b_m}^{b_{i(m-1)}}, \tag{13}$$

where subscripts $m$ and $m-1$ denote $t_m$ and $t_{m-1}$, respectively; subscript $w_{i(m-1)}$ denotes the historical w-frame fixed to the i-frame; subscript $b_{i(m-1)}$ denotes the historical b-frame fixed to the i-frame; and $\mathbf{q}_{b_{i(m-1)}}^{w_{i(m-1)}} = \mathbf{q}_{b_{m-1}}^w$ denotes the rotation of the b-frame with respect to the w-frame at $t_{m-1}$. The quaternion $\mathbf{q}_{w_{i(m-1)}}^w(t_m)$ is caused by the Earth's rotation, and its rotation vector can be expressed as

$$\boldsymbol{\phi}_{w_{i(m-1)}}^w(t_m) = -\boldsymbol{w}_{ie}^w \Delta t_{m-1,m}. \tag{14}$$

The rotation vector of the b-frame corresponding to $\mathbf{q}_{b_m}^{b_{i(m-1)}}$ can be obtained as

$$\boldsymbol{\phi}_{b_m}^{b_{i(m-1)}} \approx \Delta\boldsymbol{\theta}_m + \frac{1}{12}\Delta\boldsymbol{\theta}_{m-1} \times \Delta\boldsymbol{\theta}_m, \tag{15}$$

where the second term on the right of (15) is the second-order coning correction term [1]-[4]. The velocity update in the w-frame can be written as

$$\boldsymbol{v}_{wb_m}^w = \boldsymbol{v}_{wb_{m-1}}^w + \Delta\boldsymbol{v}_{f,m}^w + \Delta\boldsymbol{v}_{g/cor,m}^w, \tag{16}$$

where $\Delta\boldsymbol{v}_{g/cor}^{w_m}$ is the velocity increment due to gravity and the Coriolis force and $\Delta\boldsymbol{v}_{f,m}^w$ is the velocity increment due to the specific force. The latter can be expressed as follows:

$$\Delta\boldsymbol{v}_{f,m}^w = 0.5\left[\mathbf{C}_{w_{i(m-1)}}^w(t_m) + \mathbf{I}\right]\mathbf{C}_{b_{i(m-1)}}^{w_{i(m-1)}}\Delta\boldsymbol{v}_{f,m}^{b,m-1}, \tag{17}$$

$$\Delta\boldsymbol{v}_{f,m}^{b,m-1} \approx \Delta\boldsymbol{v}_{f,m}^b + 0.5\,\Delta\boldsymbol{\theta}_m \times \Delta\boldsymbol{v}_{f,m}^b + (\Delta\boldsymbol{\theta}_{m-1} \times \Delta\boldsymbol{v}_{f,m}^b + \Delta\boldsymbol{v}_{f,m-1}^b \times \Delta\boldsymbol{\theta}_m)/12, \tag{18}$$

where the second and third terms on the right hand side of (18) correspond to the rotational and sculling motions, respectively [1]-[4]. The gravity and Coriolis correction term in (16) can be computed as follows:

$$\Delta \boldsymbol{v}_{cor,m}^{w} \approx \left(\boldsymbol{g}^{w} - 2\boldsymbol{w}_{ie}^{w} \times \boldsymbol{v}_{wb_{m-1}}^{w}\right)\Delta t_{m-1,m}. \quad (19)$$

With the updated velocity, the position update in the w-frame can be simply computed as

$$\boldsymbol{p}_{wb_m}^{w} = \boldsymbol{p}_{wb_{m-1}}^{w} + 0.5\left(\boldsymbol{v}_{wb_{m-1}}^{w} + \boldsymbol{v}_{wb_m}^{w}\right)\Delta t_{m-1,m}. \quad (20)$$

### C. Preintegration Measurements

For a new epoch of IMU measurement, we use (13), (16), and (20) to update the motion states (attitude, velocity, and position). However, the IMU motion integration is dependent on the initial motion states, and thus cannot be effectively employed in an FGO framework. To avoid repeating the integration multiple times, whenever the linearized point (or bias-estimation) is changed, the preintegration theory is used to reformulate the model. To remove the dependence of the initial motion states, the b-frame fixed to the e-frame at instant $t_{k-1}$, *i.e.*, $b_{e(k-1)}$, is defined as the reference frame for the refined preintegration model.

As showed in Fig. 1, the attitude preintegration relative to $t_{k-1}$ can be computed as follows:

$$\mathbf{q}_{k-1,m}^{Pre} = \mathbf{q}_{b_m}^{b_{i(k-1)}} = \mathbf{q}_{b_{i(m-1)}}^{b_{i(k-1)}} \otimes \mathbf{q}_{b_m}^{b_{i(m-1)}}, \quad (21)$$

where $\mathbf{q}_{k-1,m}^{Pre} = \mathbf{q}_{b_m}^{b_{i(k-1)}}$ is defined as the attitude preintegration at time $t_m$, $\mathbf{q}_{b_m}^{b_{i(m-1)}}$ corresponds to the rotation vector defined in (14), and the initial state $\mathbf{q}_{k-1,k-1}^{Pre}$ is set to the identity quaternion. From (13) and (21), we can write the attitude preintegration measurement from $t_{k-1}$ to $t_k$ as

$$\mathbf{q}_{k-1,k}^{Pre} = \left(\left(\mathbf{q}_{b_k}^{w}\right)^{-1} \otimes \mathbf{q}_{w_{i(k-1)}}^{w}(t_k) \otimes \mathbf{q}_{b_{k-1}}^{w}\right)^{-1}, \quad (22)$$

where $\mathbf{q}_{w_{i(k-1)}}^{w}(t_k)$ is the Earth-rotation correction term, which can be obtained from (14). The transformation between $\mathbf{q}_{k-1,k}^{Pre}$ and $\mathbf{q}_{b_k}^{b_{e(k-1)}}$ can be expressed as follows:

$$\mathbf{q}_{b_k}^{b_{e(k-1)}} = \left(\mathbf{q}_{b_{k-1}}^{w}\right)^{-1} \otimes \mathbf{q}_{w_{i(k-1)}}^{w}(t_k) \otimes \mathbf{q}_{b_{k-1}}^{w} \otimes \mathbf{q}_{k-1,k}^{Pre}, \quad (23)$$

where the w-frame is fixed to the e-frame, and thus we have $\mathbf{q}_{b_{e(k-1)}}^{w} = \mathbf{q}_{b_{k-1}}^{w}$.

The velocity integration from $t_{k-1}$ to $t_k$ can be reformulated from (16) as

$$\boldsymbol{v}_{wb_k}^{w} = \boldsymbol{v}_{wb_{k-1}}^{w} + \mathbf{C}_{b_{e(k-1)}}^{w}\Delta\boldsymbol{v}_{k-1,k}^{Pre} + \boldsymbol{g}^{w}\Delta t_{k-1,k} - \Delta\boldsymbol{v}_{g/cor,k-1,k}^{w}. \quad (24)$$

The velocity preintegration measurement $\Delta\boldsymbol{v}_{k-1,k}^{Pre}$ in (24) is defined as follows:

$$\Delta\boldsymbol{v}_{k-1,k}^{Pre} = \int_{t_{k-1}}^{t_k} \mathbf{C}_{b(t)}^{b_{e(k-1)}} \boldsymbol{f}^{b(t)} dt$$
$$= \left(\mathbf{C}_{b_{k-1}}^{w}\right)^{T}\left(\begin{array}{c} \boldsymbol{v}_{wb_k}^{w} - \boldsymbol{v}_{wb_{k-1}}^{w} \\ -\boldsymbol{g}^{w}\Delta t_{k-1,k} + \Delta\boldsymbol{v}_{g/cor,k-1,k}^{w} \end{array}\right), \quad (25)$$

where $\mathbf{C}_{b(t)}^{b_{e(k-1)}}$ can be transformed from (23). Here, $\Delta\boldsymbol{v}_{g/cor,k-1,k}^{w}$ is the Coriolis correction term for velocity preintegration, which can be analytically computed as

$$\Delta\boldsymbol{v}_{g/cor,k-1,k}^{w} = \int_{t_{k-1}}^{t_k} 2[\boldsymbol{w}_{ie}^{w}\times]\boldsymbol{v}_{wb(t)}^{w} dt$$
$$= 2[\boldsymbol{w}_{ie}^{w}\times]\left(\boldsymbol{p}_{wb_k}^{w} - \boldsymbol{p}_{wb_{k-1}}^{w}\right). \quad (26)$$

Meanwhile, we reformulate the position integration (20) as follows:

$$\boldsymbol{p}_{wb_k}^{w} = \boldsymbol{p}_{wb_{k-1}}^{w} + \boldsymbol{v}_{wb_{k-1}}^{w}\Delta t_{k-1,k} + \mathbf{C}_{b_{e(k-1)}}^{w}\Delta\boldsymbol{p}_{k-1,k}^{Pre} + 0.5\boldsymbol{g}^{w}\Delta t_{k-1,k}^{2} - \Delta\boldsymbol{p}_{g/cor,k-1,k}^{w} \quad (27)$$

where $\Delta\boldsymbol{p}_{k-1,k}^{Pre}$ is the position preintegration measurement

$$\Delta\boldsymbol{p}_{k-1,k}^{Pre} = \iint_{t_{k-1}}^{t_k} \mathbf{C}_{b(t)}^{b_{e(k-1)}} \boldsymbol{f}^{b(t)} dt$$
$$= \left(\mathbf{C}_{b_{k-1}}^{w}\right)^{T}\left(\begin{array}{c} \boldsymbol{p}_{wb_k}^{w} - \boldsymbol{p}_{wb_{k-1}}^{w} - \boldsymbol{v}_{wb_{k-1}}^{w}\Delta t_{k-1,k} \\ -0.5\boldsymbol{g}^{w}\Delta t_{k-1,k}^{2} + \Delta\boldsymbol{p}_{g/cor,k-1,k}^{w} \end{array}\right), \quad (28)$$

where $\Delta\boldsymbol{p}_{g/cor,k-1,k}^{w}$ is the Coriolis correction term for position preintegration, which can be expressed as

$$\Delta\boldsymbol{p}_{g/cor,k-1,k}^{w} = \iint_{t_{k-1}}^{t_k} 2[\boldsymbol{w}_{ie}^{w}\times]\boldsymbol{v}_{wb(t)}^{w} dt$$
$$= 2[\boldsymbol{w}_{ie}^{w}\times]\sum_{t_m}\left[\left(\boldsymbol{p}_{wb_m}^{w} - \boldsymbol{p}_{wb_{k-1}}^{w}\right)\Delta t_{m-1,m}\right]. \quad (29)$$

From (22), (25), and (28), we can obtain the refined preintegration measurements with the Earth-rotation correction. The biases are assumed to be unchanged throughout the entire preintegration interval from $t_{k-1}$ to $t_k$. Therefore, once the bias-estimations are changed, we must update the preintegration measurements (Section III.E).

### D. Noise Propagation

In this subsection, we derive the statistic of the preintegration error state vector. The noise covariance strongly influences the MAP estimator, because the inverse noise covariance (information matrix) is used to weight the factor in the FGO [11], [13]. We define the error state vector as follows:

$$\delta\boldsymbol{z}_k = \left[\left(\delta\boldsymbol{p}_{k-1,t}^{Pre}\right)^{T} \left(\delta\boldsymbol{v}_{k-1,t}^{Pre}\right)^{T} \left(\delta\boldsymbol{\phi}_{k-1,t}^{Pre}\right)^{T} \left(\delta\boldsymbol{b}_{g_t}\right)^{T} \left(\delta\boldsymbol{b}_{a_t}\right)^{T}\right]^{T}, (30)$$

where $\delta\boldsymbol{p}_{k-1,t}^{Pre}$, $\delta\boldsymbol{v}_{k-1,t}^{Pre}$, and $\delta\boldsymbol{\phi}_{k-1,t}^{Pre}$ are the position, velocity, and attitude (in terms of the rotation vector) errors of the preintegration measurements, respectively; and $\delta\boldsymbol{b}_{g_t}$ and $\delta\boldsymbol{b}_{a_t}$ are the gyroscope and accelerometer bias errors, respectively.

In accordance with the IMU measurement model expressed in (11), the random noises $\boldsymbol{n}_g$ and $\boldsymbol{n}_a$ are modeled as Gaussian white noise processes, while the time-varying bias errors are modeled as first-order Gaussian Markov processes [3], [4]. Hence, the IMU noise model can be expressed as follows:

$$\boldsymbol{n}_g \sim \mathcal{N}(\boldsymbol{0}, \sigma_g^2\mathbf{I}),$$
$$\boldsymbol{n}_a \sim \mathcal{N}(\boldsymbol{0}, \sigma_a^2\mathbf{I}), \quad (31)$$

$$\delta\dot{\boldsymbol{b}}_{g_t} = -\frac{1}{\tau_b}\delta\boldsymbol{b}_{g_t} + \boldsymbol{n}_{b_g}, \ \boldsymbol{n}_{b_g} \sim \mathcal{N}(\boldsymbol{0}, \sigma_{b_g}^2\mathbf{I}),$$
$$\delta\dot{\boldsymbol{b}}_{a_t} = -\frac{1}{\tau_b}\delta\boldsymbol{b}_{a_t} + \boldsymbol{n}_{b_a}, \ \boldsymbol{n}_{b_a} \sim \mathcal{N}(\boldsymbol{0}, \sigma_{b_a}^2\mathbf{I}), \quad (32)$$

where $\tau_b$ and $\boldsymbol{n}_b$ are the correlation time and white noise of the first-order Gaussian Markov process, respectively.

Using the error-perturbation method in [3], [4], we can derive the continuous-time dynamics of the preintegration error

state as

$$\delta\dot{\boldsymbol{z}}_t = \boldsymbol{F}_t \delta\boldsymbol{z}_t + \boldsymbol{G}_t \boldsymbol{w}_t, \tag{33}$$

where $\boldsymbol{w}_t$ is the noise vector, which is defined as

$$\boldsymbol{w}_t = \begin{bmatrix} (\boldsymbol{n}_g)^T & (\boldsymbol{n}_a)^T & (\boldsymbol{n}_{b_g})^T & (\boldsymbol{n}_{b_a})^T \end{bmatrix}^T. \tag{34}$$

By omitting the second-order small terms, the dynamics matrix $\boldsymbol{F}_t$ is analytically expressed as follows:

$$\boldsymbol{F}_t = \begin{bmatrix} \mathbf{0} & \mathbf{I} & \mathbf{0} & \mathbf{0} & \mathbf{0} \\ \mathbf{0} & \mathbf{0} & \boldsymbol{F}_{23} & \mathbf{0} & \boldsymbol{F}_{25} \\ \mathbf{0} & \mathbf{0} & \boldsymbol{F}_{33} & -\mathbf{I} & \mathbf{0} \\ \mathbf{0} & \mathbf{0} & \mathbf{0} & -1/\tau_{b_g}\mathbf{I} & \mathbf{0} \\ \mathbf{0} & \mathbf{0} & \mathbf{0} & \mathbf{0} & -1/\tau_{b_a}\mathbf{I} \end{bmatrix}, \tag{35}$$

where the sub-matrix in $\boldsymbol{F}_t$ is defined as

$$\begin{cases} \boldsymbol{F}_{23} = -\left(\mathbf{C}_{b_{k-1}}^w\right)^T \mathbf{C}_{w_{i(k-1)}}^w(t)\mathbf{C}_{b_{k-1}}^w \hat{\mathbf{C}}_{b_t}^{b_{(k-1)}}\left[\hat{\boldsymbol{f}}^{b(t)}\times\right], \\ \boldsymbol{F}_{25} = -\left(\mathbf{C}_{b_{k-1}}^w\right)^T \mathbf{C}_{w_{i(k-1)}}^w(t)\mathbf{C}_{b_{k-1}}^w \hat{\mathbf{C}}_{b_t}^{b_{(k-1)}}, \\ \boldsymbol{F}_{33} = -[\hat{\boldsymbol{w}}_{ib(t)}^b\times]. \end{cases} \tag{36}$$

Here, the second-order terms of $\Delta t$ are also omitted, because $\Delta t$ is usually a small term. Further, $\hat{\mathbf{C}}_{b_t}^{b_{(k-1)}}$ corresponds to the attitude preintegration quaternion in (21), and $\hat{\bullet}$ denotes the estimated term, because we compensate the raw IMU measurements with the estimated IMU biases, such that

$$\begin{aligned} \hat{\boldsymbol{w}}_{ib(t)}^b &= \tilde{\boldsymbol{w}}_{ib(t)}^b - \overline{\boldsymbol{b}}_{g_{k-1}}, \\ \hat{\boldsymbol{f}}^{b(t)} &= \tilde{\boldsymbol{f}}^{b(t)} - \overline{\boldsymbol{b}}_{a_{k-1}}. \end{aligned} \tag{37}$$

Here, $\overline{\boldsymbol{b}}_{k-1}$ indicates that the biases used in the preintegration computation are constants, which is why we must incorporate bias updates (Section III.E). The noise-input mapping matrix $\boldsymbol{G}_t$ is expressed as follows:

$$\boldsymbol{G}_t = \begin{bmatrix} \mathbf{0} & \mathbf{0} & \mathbf{0} & \mathbf{0} \\ \mathbf{0} & \boldsymbol{G}_{22} & \mathbf{0} & \mathbf{0} \\ -\mathbf{I} & \mathbf{0} & \mathbf{0} & \mathbf{0} \\ \mathbf{0} & \mathbf{0} & \mathbf{I} & \mathbf{0} \\ \mathbf{0} & \mathbf{0} & \mathbf{0} & \mathbf{I} \end{bmatrix} \tag{38}$$

where the sub-matrix $\boldsymbol{G}_{22}$ is expressed as

$$\boldsymbol{G}_{22} = -\left(\mathbf{C}_{b_{k-1}}^w\right)^T \mathbf{C}_{w_{i(k-1)}}^w(t)\mathbf{C}_{b_{k-1}}^w \hat{\mathbf{C}}_{b_t}^{b_{(k-1)}}. \tag{39}$$

Because high-rate-sampled IMU data are used, the following numerical approximation can be applied to calculate the transition matrix in discrete-time form:

$$\boldsymbol{\Phi}_m = \exp\left(\boldsymbol{F}(t_{m-1})\Delta t_{m-1,m}\right) \approx \mathbf{I} + \boldsymbol{F}(t_{m-1})\Delta t_{m-1,m}. \tag{40}$$

In addition, with the continuous-time noise covariance matrix $\boldsymbol{Q}_{t_m}$, we can implement a trapezoidal integration to compute the discrete-time noise covariance matrix $\boldsymbol{Q}_m$ as follows:

$$\boldsymbol{Q}_{t_m} = \text{diag}\left(\sigma_g^2\mathbf{I}, \ \sigma_a^2\mathbf{I}, \ \sigma_{b_g}^2\mathbf{I}, \ \sigma_{b_a}^2\mathbf{I}\right), \tag{41}$$

$$\boldsymbol{Q}_m \approx 0.5\left(\boldsymbol{\Phi}_m\boldsymbol{G}_{t_m}\boldsymbol{Q}_{t_m}\boldsymbol{G}_{t_m}^T + \boldsymbol{G}_{t_m}\boldsymbol{Q}_{t_m}\boldsymbol{G}_{t_m}^T\boldsymbol{\Phi}_m^T\right)\Delta t_{m-1,m}, \tag{42}$$

where $\boldsymbol{Q}_{t_m}$ is the covariance matrix of the noise vector $\boldsymbol{w}_t$ in (34). Then, the covariance matrix $\boldsymbol{\Sigma}_{k-1,m}^{Pre}$ can be propagated from the initial covariance $\boldsymbol{\Sigma}_{k-1,k-1}^{Pre} = \mathbf{0}$ as follows:

$$\boldsymbol{\Sigma}_{k-1,m}^{Pre} = \boldsymbol{\Phi}_m\boldsymbol{\Sigma}_{k-1,m-1}^{Pre}\boldsymbol{\Phi}_m^T + \boldsymbol{Q}_m. \tag{43}$$

The first-order Jacobian matrix can also be propagated recursively with the initial Jacobian $\mathbf{J}_{k-1,k-1} = \mathbf{I}$, as

$$\mathbf{J}_{k-1,m} = \boldsymbol{\Phi}_m\mathbf{J}_{k-1,m-1}. \tag{44}$$

Using recursive formulation in (43) and (44), we can obtain the $\boldsymbol{\Sigma}_{k-1,k}^{Pre}$ and $\mathbf{J}_{k-1,k}$, which span the entire preintegration interval.

### E. Bias Updates

In Sections III.B and III.C, the biases are assumed unchanged throughout the entire preintegration interval (see (37)). During the FGO solution solving process, if the bias estimations are changed, we must incorporate bias updates, in order to update the preintegration measurements accordingly. The altered biases can be computed as follows:

$$\begin{aligned} \delta\boldsymbol{b}_g &= \boldsymbol{b}_{g_{k-1}} - \overline{\boldsymbol{b}}_{g_{k-1}}, \\ \delta\boldsymbol{b}_a &= \boldsymbol{b}_{a_{k-1}} - \overline{\boldsymbol{b}}_{a_{k-1}}, \end{aligned} \tag{45}$$

where $\overline{\boldsymbol{b}}_{g_{k-1}}$ and $\overline{\boldsymbol{b}}_{a_{k-1}}$ denote the constant biases in (37), while $\boldsymbol{b}_{g_{k-1}}$ and $\boldsymbol{b}_{a_{k-1}}$ denote the newly estimated biases. Then, we can update the preintegration measurements computed from (22), (25), and (28) using first-order expansions

$$\Delta\hat{\boldsymbol{p}}_{k-1,k}(\boldsymbol{b}) \approx \Delta\hat{\boldsymbol{p}}_{k-1,k}^{Pre}(\overline{\boldsymbol{b}}) + \mathbf{J}_{k-1,k}^{\boldsymbol{p},\boldsymbol{b}_g}\delta\boldsymbol{b}_g + \mathbf{J}_{k-1,k}^{\boldsymbol{p},\boldsymbol{b}_a}\delta\boldsymbol{b}_a,$$

$$\Delta\hat{\boldsymbol{v}}_{k-1,k}^{Pre}(\boldsymbol{b}) \approx \Delta\hat{\boldsymbol{v}}_{k-1,k}^{Pre}(\overline{\boldsymbol{b}}) + \mathbf{J}_{k-1,k}^{\boldsymbol{v},\boldsymbol{b}_g}\delta\boldsymbol{b}_g + \mathbf{J}_{k-1,k}^{\boldsymbol{v},\boldsymbol{b}_a}\delta\boldsymbol{b}_a, \tag{46}$$

$$\hat{\mathbf{q}}_{k-1,k}^{Pre}(\boldsymbol{b}) \approx \hat{\mathbf{q}}_{k-1,k}^{Pre}(\overline{\boldsymbol{b}}) \otimes \begin{bmatrix} 1 \\ \mathbf{J}_{k-1,k}^{\phi,\boldsymbol{b}_g}\delta\boldsymbol{b}_g \end{bmatrix},$$

where the left hand sides of (46) are the preintegration measurements updated by the newly estimated biases in (45); the first term of each formulation on the right side of (46) is the preintegration measurements computed using the constant biases in (45); and $\mathbf{J}_{k-1,k}^{\boldsymbol{p},\boldsymbol{b}_g}$ is the sub-matrix of $\mathbf{J}_{k-1,k}$, the location of which corresponds to $\delta\boldsymbol{p}_{k-1,k}^{Pre}/\delta\boldsymbol{b}_{g_{k-1}}$ (the same definition applies to $\mathbf{J}_{k-1,k}^{\boldsymbol{p},\boldsymbol{b}_a}$, $\mathbf{J}_{k-1,k}^{\boldsymbol{v},\boldsymbol{b}_g}$, $\mathbf{J}_{k-1,k}^{\boldsymbol{v},\boldsymbol{b}_a}$, and $\mathbf{J}_{k-1,k}^{\phi,\boldsymbol{b}_g}$).

Finally, we obtain an accurate IMU preintegration model, which is derived from the classic INS kinematic model while incorporating the Earth's rotation. In addition, the noise propagation procedure of the preintegration measurement is elaborately described.

## IV. FGO-BASED GNSS/INS INTEGRATION

In this section, we construct an FGO-based GNSS/INS integrated navigation system using the refined IMU preintegration model described in the previous section, so as to quantitatively evaluate the accuracy of preintegration model. We first formulate the sliding-windows optimizer, followed by the IMU preintegration factor and the GNSS positioning factor. Finally, we briefly describe the marginalization.

### A. Formulation

An FGO-based sliding-window optimizer is utilized to process the IMU preintegration and GNSS positioning measurements, as depicted in Fig. 2. The IMU states in the sliding window are defined as $\boldsymbol{X} = [\boldsymbol{x}_0, \ \boldsymbol{x}_1, \ ..., \ \boldsymbol{x}_n]$, where $\boldsymbol{x}_k$ is expressed as

$$\boldsymbol{x}_k = \left[\left(\boldsymbol{p}_{\mathrm{wb}_k}^{\mathrm{w}}\right)^T, \left(\mathbf{q}_{\mathrm{b}_k}^{\mathrm{w}}\right)^T, \left(\boldsymbol{v}_{\mathrm{wb}_k}^{\mathrm{w}}\right)^T, \left(\boldsymbol{b}_{g_k}\right)^T, \left(\boldsymbol{b}_{a_k}\right)^T\right]^T. \quad (47)$$

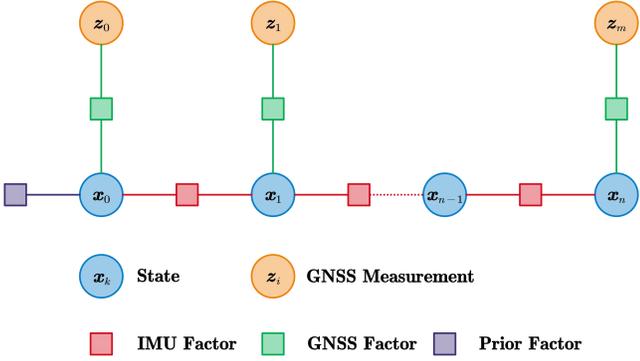

Fig. 2. Illustration of FGO-based GNSS/INS integration navigation system.

Here, the state $\boldsymbol{x}_k$ includes the position, velocity, and attitude of the IMU in the w-frame, along with the gyroscope and accelerometer biases, where $k \in [0, n]$ and $n$ is the sliding-window size. We add an IMU preintegration factor into the sliding window at each GNSS second, even though the GNSS positioning factor may be absent. In other words, the interval of each preintegration factor is 1 s.

We minimize the sum of the prior and the Mahalanobis norm of all measurement residuals to obtain the following MAP estimation:

$$\min_{\boldsymbol{X}} \left\{ \begin{array}{l} \|\mathbf{r}_p - \mathbf{H}_p \boldsymbol{X}\|^2 + \sum_{k \in \{1,\, n\}} \left\|\mathbf{r}_{Pre}\left(\hat{\boldsymbol{z}}_{k-1,k}^{Pre}, \, \boldsymbol{X}\right)\right\|_{\Sigma_{k-1,k}^{Pre}}^2 \\ + \sum_{i \in [0,\, m]} \left\|\mathbf{r}_{GNSS}\left(\hat{\boldsymbol{z}}_i^{GNSS}, \, \boldsymbol{X}\right)\right\|_{\Sigma_i^{GNSS}}^2 \end{array} \right\}, (48)$$

where $\mathbf{r}_{Pre}$ and $\mathbf{r}_{GNSS}$ are the residuals for the IMU preintegration factors and GNSS positioning factors, respectively; $m+1$ $(m \leq n)$ is the number of GNSS positioning factors, and $\{\mathbf{r}_p, \mathbf{H}_p\}$ represents the prior information from marginalization (Section IV.D). The Ceres solver [25] is used for this nonlinear problem.

### B. IMU Preintegration Factor

In accordance with the preintegration measurements in (22), (25), and (28), and the bias-updated preintegration measurements in (46), we can compute the residual of the IMU preintegration factor as

$$\mathbf{r}_{Pre}\left(\hat{\boldsymbol{z}}_{k-1,k}^{Pre}, \, \boldsymbol{X}\right) = \begin{bmatrix} \Delta\boldsymbol{p}_{k-1,k}^{Pre} - \Delta\hat{\boldsymbol{p}}_{k-1,k}^{Pre} \\ \Delta\boldsymbol{v}_{k-1,k}^{Pre} - \Delta\hat{\boldsymbol{v}}_{k-1,k}^{Pre} \\ 2\left[\left(\mathbf{q}_{\mathrm{b}_k}^{\mathrm{w}}\right)^{-1} \otimes \mathbf{q}_{\mathrm{w}(k-1)}^{\mathrm{w}}(t_k) \otimes \mathbf{q}_{\mathrm{b}_{k-1}}^{\mathrm{w}} \otimes \hat{\mathbf{q}}_{k-1,k}^{Pre}\right]_v \\ \boldsymbol{b}_{g_k} - \boldsymbol{b}_{g_{k-1}} \\ \boldsymbol{b}_{a_k} - \boldsymbol{b}_{a_{k-1}} \end{bmatrix}, (49)$$

where $2[\cdot]_v$ is the inverse algorithm of (3) to extract the (small-angle) rotation vector of a quaternion. The gyroscope and accelerometer biases are also included in the residual terms for online correction.

### C. GNSS Positioning Factor

The positioning result in geodetic coordinates $\hat{\boldsymbol{p}}_{GNSS,i}^e$ and its covariance $\boldsymbol{\Sigma}_i^{GNSS}$ can be obtained from the GNSS receiver. The geodetic coordinates can then be converted to the local w-frame as $\hat{\boldsymbol{p}}_{GNSS,i}^{\mathrm{w}}$ [3], [4]. Hence, the residual of the GNSS positioning factor can be expressed as follows:

$$\mathbf{r}_{GNSS}\left(\hat{\boldsymbol{z}}_i^{GNSS}, \, \boldsymbol{X}\right) = \boldsymbol{p}_{\mathrm{wb}_i}^{\mathrm{w}} + \mathbf{C}_{\mathrm{b}}^{\mathrm{w}} \boldsymbol{l}_{GNSS}^{\mathrm{b}} - \hat{\boldsymbol{p}}_{GNSS,i}^{\mathrm{w}}, \quad (50)$$

where $\boldsymbol{l}_{GNSS}^{\mathrm{b}}$ is the GNSS antenna lever-arm expressed in the IMU b-frame, and covariance $\boldsymbol{\Sigma}_i^{GNSS}$ corresponds to the NED direction of the w-frame (Section II.B).

### D. Marginalization

We incorporate marginalization to bound the computational complexity of the sliding-window optimizer. When the IMU preintegration factor exceeds the threshold (the sliding-window size), we marginalize out the oldest IMU state, and convert the IMU preintegration measurement and GNSS positioning measurement corresponding to the marginalized state into a prior factor. For more details on the marginalization in sliding-window optimization, readers may refer to [26].

The refined IMU preintegration model is incorporated into an FGO-based GNSS/INS integrated navigation system and the residuals of the IMU preintegration factor and GNSS positioning factor are computed as analytical expressions. The marginalization approach is also adopted to reduce the computational cost.

## V. Experimental Results

This section reports verification and evaluation of the refined IMU preintegration model, which is incorporated into the FGO-based GNSS/INS integration system described in Section IV. A classic EKF-based GNSS/INS integrated navigation system using high-accuracy INS mechanization [1]-[4] was adopted as a baseline, as this system constitutes a benchmark of the potential system accuracies (especially the potential INS accuracy). The simulated GNSS outage [24] was used to quantitatively evaluate the accuracy of the integrated systems. Below, the experiment setup is described first; then, the results and accompanying discussion are presented.

### A. Vehicle Experiment Setup

Three vehicle tests were conducted in an open-sky area, having durations of 2325, 1617, and 2333 s, respectively. Throughout the entire travel period, the average vehicle speed was approximately 10 m/s. Allowing at least 500 s for navigation-system initialization, we intentionally blocked the GNSS positioning for 60 s at 150-s intervals in post-processing. Each test sequence was processed twice with different first-outage start times. We finally obtained 61 outages in total, and the statistical results for the maximum position-drift error during each simulated GNSS outage were adopted to evaluate the accuracies of the integrated systems. Note that the attitude drift was not considered here because, the position drift is a sensitive indicator of the inherent accuracy of the entire navigation system, while the attitude drift is not [24]. Further details on simulated GNSS outage are available in [24]. Specifically, the root mean square errors (RMSEs) of the maximum horizontal- and vertical-drift errors in each GNSS outage were used. A typical position-drift error curve for the simulated GNSS outage is depicted in Fig. 3.

TABLE I
BIAS INSTABILITY PARAMETERS OF MEMS IMUS

| IMU | ICM20602 (InvenSense) | ADIS16460 (ADI) | ADIS16465 (ADI) | HGuide-i300 (Honeywell) |
|---|---|---|---|---|
| Gyroscope Bias Instability (°/h 1σ) | 10.0 | 8.0 | 2.0 | 3.0 |

The gyroscope bias instability parameter for ICM20602 was obtained from the tested Allan variance curve; the ADIS16460, ADIS16465, and HGuide-i300 parameters were obtained from the corresponding device data sheets.

TABLE II
PROCESSING-MODE CONFIGURATIONS

| Mode | Type | Description |
|---|---|---|
| M0 (EKF-based) | EKF | A classic GNSS/INS integrated navigation algorithm based on EKF (used as the baseline). |
| M1 (Refined) | FGO | A mode incorporating the refined preintegration considering the Earth's rotation, as introduced in Section III. |
| M2 (Rough) | FGO | A mode incorporating a rough preintegration that lacks the Earth's rotation correction. |

The FGO-based integrations for M1 and M2 are described in Section IV.

Four different MEMS IMUs were employed in the tests: ICM20602, a consumer-grade MEMS chip; and ADIS16460, ADIS16465, and HGuide-i300, which are industrial-grade MEMS modules. The ground-truth was obtained from a navigation-grade GNSS/INS integrated navigation system. All sensors were attached to the same vehicle and sampled simultaneously at the same sample rate of 200 Hz. The main parameters of the four MEMS IMUs are listed in Table I. Note that the magnitudes of the gyroscope-bias instabilities in Table I are all less than the Earth's rotation rate of 15 °/h, which indicates the necessity of considering the Earth's rotation in IMU preintegration; otherwise, the gyroscope precision is wasted. The GNSS positioning results used in the tests were derived from Post-Processed Kinematic (PPK) [3] technology with centimeter-level accuracy.

Three processing modes were implemented for the tests: EKF-based integration and refined and rough FGO-based GNSS/INS integration, labeled M0, M1, and M2, respectively (Table II). The EKF-based integration was treated as the baseline because its accuracy has been well proven to constitute a benchmark for the potential INS accuracy [3], [4], as noted above. We conducted dedicated experiments to evaluate the effect of the sliding-window size $n$; hence, we found that this factor had little impact on the accuracy of the FGO-based GNSS/INS integration systems in this experiment. Consequently, the sliding-window sizes of both M1 and M2 were set to 20 s to bound the computation complexity.

Parameter tuning was conducted to mitigate the impacts of the IMU noise parameters. Note that these parameters are modeled in (31) and (32) for both EKF-based and FGO-based integration. The correction-times $\tau_{b_g}$ and $\tau_{b_a}$ were set to 1 h for all four MEMS IMUs without tuning, based on our previous experience. Batch processes based on grid searching were implemented to obtain the optimal parameters, by minimizing the RMSEs of the position-drift errors.

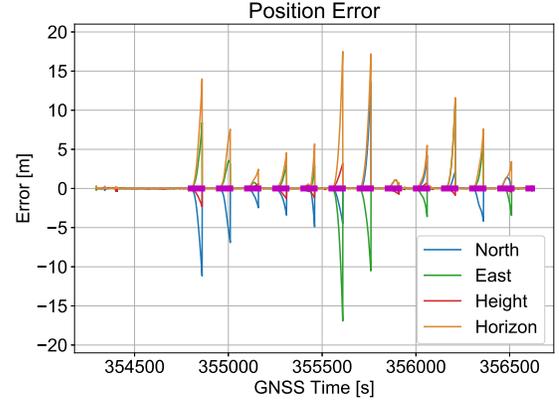

Fig. 3. Position-drift error curve for simulated GNSS outages.

TABLE III
RMSES OF HORIZONTAL AND VERTICAL DRIFTS IN METERS

| IMU | M0 (EKF-based) | | M1 (Refined) | | M2 (Rough) | |
|---|---|---|---|---|---|---|
| | Hor | Ver | Hor | Ver | Hor | Ver |
| ICM20602 | 45.10 | 12.89 | 45.30 | 12.89 | 51.03 | 13.09 |
| ADIS16460 | 20.03 | 2.30 | 20.15 | 2.26 | 29.19 | 2.43 |
| ADIS16465 | 9.13 | 0.97 | 9.14 | 0.96 | 22.43 | 1.50 |
| HGuide-i300 | 7.68 | 2.15 | 7.71 | 2.14 | 23.09 | 2.98 |

Here, "Hor" and "Ver" denote the horizontal and vertical drifts, respectively.

TABLE IV
HORIZONTAL-DRIFT DIFFERENCES BETWEEN M0 (EKF-BASED) AND M2 (ROUGH)

| IMU | ICM20602 | ADIS16460 | ADIS16465 | HGuide-i300 |
|---|---|---|---|---|
| Difference (m) | 5.93 | 9.16 | 13.3 | 15.41 |
| Percentage (%) | 13.14% | 45.73% | 145.67% | 200.65% |

The difference is defined as M2-M0, and the percentage is defined as (M2-M0)/M0*100%.

### B. Results and Discussions

Following parameter tuning, we obtained the RMSEs of the position-drift errors for the four MEMS IMUs. As apparent from Table III, the RMSEs for M0 and M1 were almost identical for all four IMUs; however, the RMSEs for M2 were much larger than those of the other two modes. These results demonstrate that the refined preintegration model can achieve the same accuracy as classic high-accuracy INS once the Earth's rotation has been properly considered; however, the accuracy may degrade significantly without Earth-rotation correction. From the horizontal-drift RMSEs for M0, HGuide-i300 and ICM20602 were the best- and worst-performing of the four IMUs considered in this experiment, respectively. However, the horizontal-drift RMSE obtained for HGuide-i300 and M2 was even larger than that for ADIS16465, illustrating that the accuracy degradation in M2 was much larger for the better-performing IMU.

We also computed the horizontal-drift differences between M0 and M2 to quantitatively evaluate the impact of the Earth-rotation correction on the preintegration. As apparent from Table IV, the better the IMU performance, the larger the horizontal-drift difference; this result corresponds to the previous conclusion. Without correction for the Earth's rotation, the horizontal-drift error increased significantly for FGO-based integration, as indicated by the rough preintegration result (M2); thus, the IMU precision was wasted. Specifically, the horizontal-drift error increased by more than 200% for HGuide-i300, an industrial-grade MEMS module, and even by more than 10% for ICM20602, the consumer-grade MEMS chip. Thus, we can conclude that Earth-rotation correction is the major factor influencing preintegration accuracy.

## VI. CONCLUSIONS

In this study, we explored the accuracy potential of IMU preintegration in FGO by considering the Earth's rotation. A refined preintegration model was constructed and quantitatively evaluated for the case of GNSS/INS integration. The experiment results obtained in this work indicate that the refined IMU preintegration model with Earth-rotation correction significantly improved the accuracy of FGO-based GNSS/INS integration to the level of the EKF-based baseline.

The results of this experiment constitute solid evidence that existing IMU preintegration models are not sufficiently precise and must be improved, even for consumer-grade MEMS IMUs. This finding should also transfer to other cases involving FGO frameworks and incorporating IMU preintegration, such as VINSs.